%% file: main.tex
\documentclass{article}
\usepackage{iclr2027_conference,times}
\input{math_commands.tex}

\usepackage{microtype}
\usepackage{booktabs}
\usepackage{enumitem}
\usepackage{graphicx}
\usepackage{wrapfig}
\usepackage{needspace}
\usepackage{capt-of}
\usepackage{tikz}
\usetikzlibrary{arrows.meta}
\usepackage{float}
\usepackage{placeins}
\usepackage{amsmath,amssymb}
\usepackage{xcolor}
\usepackage{colortbl}
\usepackage{hyperref}
\hypersetup{hidelinks}
\usepackage{url}


\definecolor{methodblue}{HTML}{2F6BFF}
\definecolor{questionboxfill}{HTML}{F2EFEB}
\definecolor{questionboxborder}{HTML}{B8AEA4}
\definecolor{questionboxtext}{HTML}{4F4A46}
\newcommand{\method}{\textsc{CacheLink}}

\title{CacheLink: Functional KV Transport for Zero-Prefill Cross-Model Context Handoff}

\author{%
  {\small Haomiao Qiu$^{1,2,*}$ \; Yining Tang$^{3,*}$ \; Zijia Song$^{1}$ \; Ruizi Han$^{1}$ \; Ziyue Qiao$^{2,\dagger}$ \; Miao Zhang$^{1,\dagger}$}\\[0.4em]
  {\footnotesize $^{1}$Harbin Institute of Technology, Shenzhen \quad $^{2}$Great Bay University \quad $^{3}$Zhejiang University}\\[0.2em]
  {\footnotesize \href{mailto:24B951058@stu.hit.edu.cn}{\texttt{24B951058@stu.hit.edu.cn}}}
}

\iclrfinalcopy 
\begin{document}
\maketitle
\lhead{Preprint}
\begingroup
\renewcommand{\thefootnote}{\fnsymbol{footnote}}
\footnotetext[0]{\textsuperscript{*} Equal contribution.\qquad \textsuperscript{$\dagger$} Corresponding authors.}
\endgroup

\begin{abstract}
KV caches make long-context serving efficient by reusing past computation, but model switching breaks this reuse and forces the receiver to prefill the context again. We study \emph{zero-prefill cross-model context handoff}: translating a sender cache into a receiver-compatible state so that the receiver can continue directly without prefilling same context. 
Existing affine transport methods show that receiver KV states can be substantially reconstructed from sender caches. However, through extensive analyses, we find that reconstruction fidelity is a poor proxy for receiver utility. Similar K/V reconstruction errors can lead to substantially different attention patterns, predictive distributions, and downstream accuracy, because conventional reconstruction objectives overlook the query-dependent functional importance of K and V in receiver attention.
We thus introduce \method{}, which learns a coupled K/V correction through the receiver behavior directly. Across four independently trained sender--receiver pairs and seven evaluation panels, \method{} improves the equal-weight mean over all methods for every pair. In matched diagnostics, \method{} changes K/V squared error by less than 0.4\% while substantially changing receiver attention and next-token behavior. At 32K context, it reduces handoff latency by up to $7.27\times$ relative to receiver re-prefill. Our findings verify that reconstruction fidelity alone is insufficient and that preserving the receiver computation induced by the transported cache is important for effective cross-model KV transport. 
\end{abstract}

\input{sections/introduction}
\input{sections/related_work}
\input{sections/problem_setup}
\input{sections/motivation}

\input{sections/method}
\input{sections/experiments}
\input{sections/conclusion}

\subsubsection*{Reproducibility statement}
Appendix~\ref{app:protocol} specifies each model pair's cache convention, mapper construction, prompts, decoding, and scoring. The anonymized source artifact contains the selected training and evaluation code but does not bundle model weights, Ridge fits, benchmark panels, predictions, or run logs. The manuscript artifact provides the fixed-panel result tables and, for the matched diagnostic, sanitized row identifiers, panel hashes, per-example numeric metrics, and aggregation code; all required base models, checkpoints, and evaluation inputs must be prepared separately as described in the appendix.

\subsubsection*{AI use statement}
Generative AI tools were used to assist with literature search, experiment orchestration, plotting code, and language editing. All numerical claims are derived from stored experiment artifacts; the authors reviewed the generated text and code, checked citations against primary sources, and take responsibility for the final content.

\bibliography{references}
\bibliographystyle{iclr2027_conference}

\appendix
\input{sections/appendix}

\end{document}

%% file: math_commands.tex
\usepackage{amsmath,amsfonts,bm}

\def\eqref#1{equation~\ref{#1}}

\def\1{\bm{1}}

\def\vx{{\bm{x}}}

\DeclareMathAlphabet{\mathsfit}{\encodingdefault}{\sfdefault}{m}{sl}
\SetMathAlphabet{\mathsfit}{bold}{\encodingdefault}{\sfdefault}{bx}{n}



%% file: sections/introduction.tex
\section{Introduction}
\label{sec:introduction}
\begin{figure*}[!t]
    \centering
    \includegraphics[width=\textwidth]{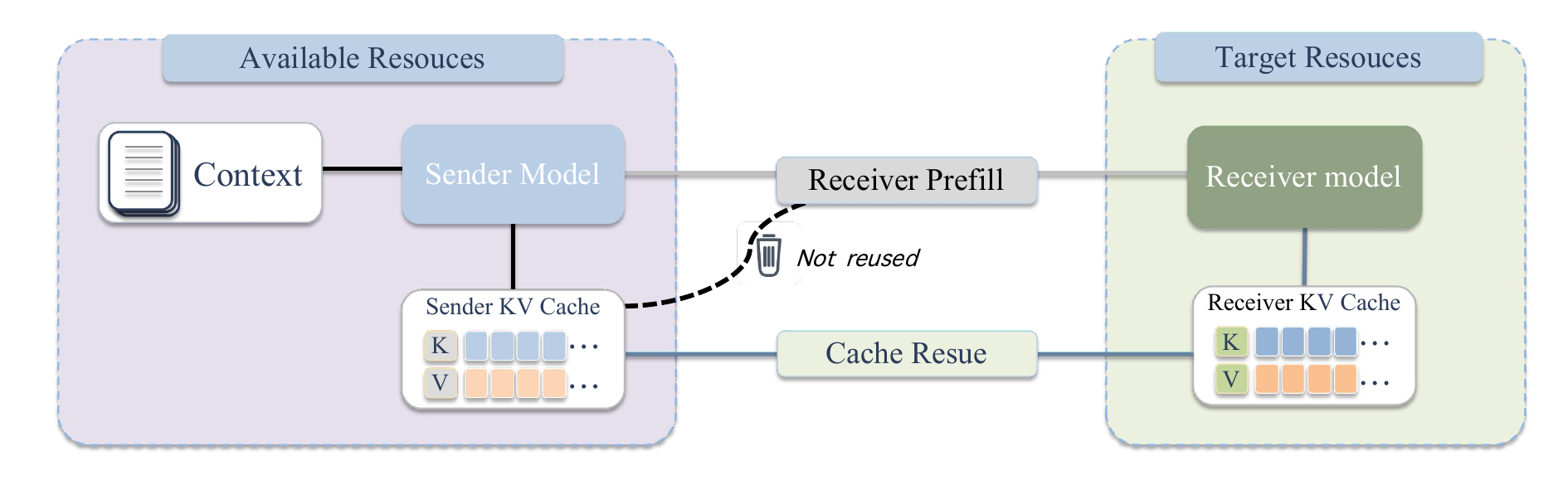}
    \caption{\textbf{A model switch breaks native KV-cache reuse.}
    The sender has already processed the accumulated context and produced a reusable cache. Native switching discards that computation and rebuilds the receiver cache by prefilling the same prefix again. Cross-model handoff translates the sender cache into a receiver-compatible state, allowing the receiver to continue without replaying the transferred context.}
    \label{fig:overview}
\end{figure*}

KV caches reduce redundant context computation and improve LLM serving efficiency, especially for long-running agents. Yet, when switching models, the cached states cannot be directly transferred to and reused by the new model, forcing it to prefill the same context again.
Besides, Routers and cascades increasingly move requests across models to trade inference cost for capability~\citep{dohan2022cascades,chen2024frugalgpt,aggarwal2024automix,ong2025routellm,hu2024routerbench,zhang2026evoroute}. A model switch therefore creates a recurring systems cost: the sender may have already processed tens of thousands of useful tokens, but the receiver cannot directly reuse its KV cache and must prefill the same context again. Same-model serving systems avoid this repeated work through prefix-KV reuse~\citep{kwon2023vllm,gim2024promptcache,srivatsa2025preble,zheng2024sglang,qin2025mooncake}. We study \emph{zero-prefill cross-model context handoff}, where the sender processes the reusable prefix once, its KV cache is translated into the receiver's cache space, and the receiver continues without replaying the transferred prefix (Figure~\ref{fig:overview}).

Recent cross-model state-transfer work shows that related models admit useful internal alignments~\citep{chen2025stitching,zhao2025iam,fu2026c2c,dery2026latent}. For KV transport, per-head Ridge regression reconstructs a substantial fraction of the receiver cache from sender states~\citep{heo2026crossmodel}, and CacheBridge further improves affine mapping across models~\citep{qu2026cachebridge}. These results establish reconstruction as a strong starting point. They do not establish reconstruction fidelity as the right objective. Receiver queries consume the translated K and V jointly through attention, and we find that caches with similar K/V error can induce substantially different attention, predictive distributions, and task accuracy.

This reconstruction--utility mismatch motivates \method{}. \method{}  learns a coupled K/V correction, and trains the transported state through the frozen receiver with functional supervision. Ridge supplies the cross-model structure already captured by affine prediction~\citep{heo2026crossmodel}; the functional objective directs subsequent correction toward the behavior that the receiver actually uses. Deployment remains zero-prefill: the receiver processes only the new query and generated tokens.

Across four independently trained sender$\rightarrow$receiver pairs and seven fixed evaluation panels, \method{} achieves a higher equal-weight mean than both Ridge and the native sender for every pair and outperforms Ridge in 26 of 28 pair--panel comparisons. Matched diagnostics sharpen the mechanism: receiver attention and next-token behavior change substantially while K/V squared error changes by less than 0.4\%. At a 32K reusable prefix, \method{} reduces handoff latency by up to $7.27\times$ relative to receiver re-prefill.

Our main contributions are:
\begin{itemize}
    \item \textbf{Problem and finding.} We formulate zero-prefill cross-model context handoff and conduct extensive diagnostic analyses of cross-model KV reuse. These analyses show that KV reconstruction fidelity does not determine downstream receiver utility: caches with similar reconstruction error can induce markedly different receiver behavior.

    \item \textbf{Method.} We introduce \method{}, which learns a coupled K/V correction through functional supervision from a frozen receiver. By optimizing the computation induced by the transported cache, \method{} targets receiver behavior directly instead of focusing solely on accurate tensor reconstruction.

    \item \textbf{Evaluation.} Across four sender$\rightarrow$receiver model pairs and seven datasets, \method{} consistently outperforms Ridge while preserving the latency advantage of zero-prefill handoff, achieving up to $7.27\times$ lower handoff latency than receiver re-prefill at 32K context.
\end{itemize}

%% file: sections/related_work.tex
\section{Related work}
\label{sec:related}

\paragraph{Cross-model representation and state transfer.}
Representation-similarity analyses show that independently trained models often share substantial geometric structure~\citep{kornblith2019cka,huh2024platonic}, while model-stitching methods exploit this structure by learning mappings between intermediate states~\citep{bansal2021stitching,chen2025stitching}. Recent LLM systems extend this idea to runtime transfer through attention mapping, KV projection, shared latent spaces, selective recomputation, or coordinated prefill~\citep{zhao2025iam,fu2026c2c,dery2026latent,liu2026droidspeak,woo2026icarus,woo2026prefillshare}.

\paragraph{Cross-model KV transport.}
KV caches are especially attractive transfer targets because they contain the state needed to continue autoregressive decoding after a prefix has been processed. Recent work shows that sender KV states can predict a substantial fraction of receiver KV structure using simple affine mappings~\citep{heo2026crossmodel}, and CacheBridge develops related cross-model cache mappings for efficient reuse~\citep{qu2026cachebridge}. Our experiments build on this reconstruction basis: because the receiver consumes K and V jointly through attention, transported-cache utility depends on the computation it induces, not only on elementwise agreement with native tensors. CacheLink therefore starts from an affine receiver-cache estimate and learns a coupled correction through receiver-level functional supervision.

\paragraph{KV reuse, serving, and model routing.}
Within a single model, modern serving systems preserve prefill work through prefix caching, shared-prefix scheduling, disaggregated serving, and selective cache reuse across repeated or overlapping requests~\citep{gim2024promptcache,zheng2024sglang,srivatsa2025preble,qin2025mooncake,yao2025cacheblend}. Complementary work reduces the memory and bandwidth cost of an already-compatible cache through quantization, token selection, and cache compression~\citep{liu2024kivi,tang2024quest,li2024snapkv}. At the same time, routers and cascades increasingly move requests among models with different capability and cost profiles~\citep{dohan2022cascades,chen2024frugalgpt,aggarwal2024automix,ong2025routellm,zhang2026evoroute}. These two trends meet in long-running workloads: context may be reusable, yet the next model selected by the router cannot directly consume the previous model's KV state. Cross-model cache transport turns that already-computed prefix state into something that can survive a model switch, allowing routing and context reuse to coexist without replaying the reusable prefix.

%% file: sections/problem_setup.tex
\section{Problem setup}
\label{sec:problem_setup}

\subsection{Zero-prefill cross-model handoff}
\label{sec:handoff_setting}

Let a sender model $S$ process a reusable prefix $x_{1:n}$ and produce
\begin{align}
    C^S(x_{1:n})
    &=
    \left\{
    K^S_l(x_{1:n}),V^S_l(x_{1:n})
    \right\}_{l=1}^{L_S},
    &\qquad
    C^T(x_{1:n})
    &=
    \left\{
    K^T_l(x_{1:n}),V^T_l(x_{1:n})
    \right\}_{l=1}^{L_T}.
\end{align}
A receiver model $T$ would ordinarily prefill the same prefix to construct $C^T(x_{1:n})$. We instead learn :
\begin{equation}
    F:\; C^S(x_{1:n}) \rightarrow \widehat C^T(x_{1:n}),
\end{equation}
 as a transport and inject $\widehat C^T$ directly into the receiver. Under the strict zero-prefill constraint used throughout this paper, deployment performs no target-prefix prefill, sparse replay, or receiver-side reconstruction pass over $x_{1:n}$. The complete receiver cache is produced from the sender cache alone.

The transported cache is the state consumed by future receiver queries. At each attention layer, K determines how a query routes probability mass across cached positions, while V determines the information returned along those routes. Handoff quality therefore depends on the receiver computation induced by the paired K/V state.

\subsection{Reconstruction baseline}
\label{sec:ridge_formulation}

The Ridge baseline treats handoff as supervised reconstruction of the native receiver cache~\citep{heo2026crossmodel}. For each target layer $l$, let $\mathcal{S}_l$ contain the $k$ sender layers selected by held-out predictability. After removing source RoPE from keys, selected states are concatenated at each token position,
\begin{equation}
    \vx^K_{l,t}
    =
    \operatorname{concat}_{j\in\mathcal{S}_l} K^S_{j,t},
    \qquad
    \vx^V_{l,t}
    =
    \operatorname{concat}_{j\in\mathcal{S}_l} V^S_{j,t}.
\end{equation}
Independent affine regressors predict each target-layer KV head,
\begin{equation}
    \begin{aligned}
    \widehat K^T_{l,h,t}
    &=
    \vx^K_{l,t}W^K_{l,h}+b^K_{l,h},
    \quad
    \widehat V^T_{l,h,t}
    &=
    \vx^V_{l,t}W^V_{l,h}+b^V_{l,h}.
    \end{aligned}
\end{equation}
\begin{equation}
    W^\star
    =
    \arg\min_W
    \left(
    \lVert XW-Y\rVert_F^2
    +
    \lambda\lVert W\rVert_F^2
    \right).
\end{equation}
Target RoPE is reapplied to $\widehat K^T$ before cache injection. Our matched baseline uses $k=8$ and the published $500\times1024$ FineWeb-Edu calibration protocol.
Ridge reconstructs a substantial fraction of the receiver KV state and provides a strong structural baseline for cross-model handoff.

\begin{center}
\setlength{\fboxsep}{6pt}%
\setlength{\fboxrule}{0.7pt}%
\fcolorbox{questionboxborder}{questionboxfill}{%
    \parbox{0.91\linewidth}{%
        \centering
        \color{questionboxtext}
        \textbf{\emph{Does lower KV reconstruction error produce a more useful receiver cache?}}
    }%
}
\end{center}

%% file: sections/motivation.tex
\section{Reconstruction fidelity does not determine receiver utility}
\label{sec:motivation}

\subsection{Reconstruction fidelity does not predict handoff utility}
\label{sec:ridge_gap}

Ridge reconstructs substantial receiver KV structure from the sender cache, yet its downstream utility varies sharply across tasks. To test whether reconstruction error explains this variation, we use a targeted Qwen3-14B$\rightarrow$32B diagnostic; the full seven-panel comparison across all four sender--receiver pairs appears in Appendix Figure~\ref{fig:ridge_short_long_four_pairs}.

\begin{figure*}[!t]
    \centering
    \includegraphics[width=\textwidth]{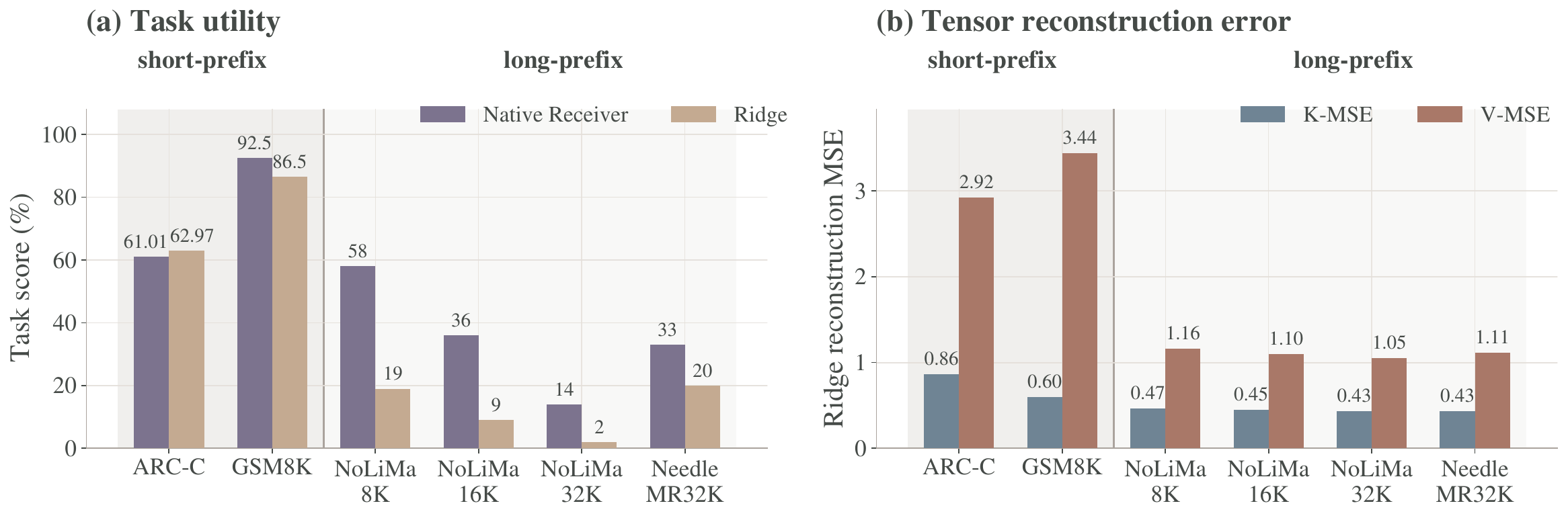}
    \caption{\textbf{Long-prefix utility degrades even as KV reconstruction error decreases.}
    Both panels use the same dataset order and the same Qwen3-14B$\rightarrow$32B diagnostic. (a) Native and Ridge task scores. (b) Ridge K/V reconstruction MSE. Across NoLiMa 8K--32K, reconstruction error decreases while task utility falls sharply.}
    \label{fig:ridge_utility_mse}
\end{figure*}

Figure~\ref{fig:ridge_utility_mse} compares ARC-C and GSM8K with NoLiMa 8K/16K/32K and NeedleBench MR32K. Ridge remains relatively close to the native receiver on the short-prefix tasks, while its NoLiMa score drops from 19\% to 9\% to 2\% as the reusable prefix grows from 8K to 16K to 32K; the native receiver scores 58\%, 36\%, and 14\%. Over the same sweep, K-MSE decreases from 0.465 to 0.433 and V-MSE decreases from 1.161 to 1.054. Utility therefore degrades while reconstruction error improves.

\FloatBarrier

\textbf{Observation 1.} Lower K/V reconstruction error does not imply higher receiver utility.

\subsection{Receiver behavior can change without better reconstruction}
\label{sec:mse_mismatch}
\label{sec:fixed_panel_diagnostics}

The receiver consumes a translated cache through attention. For a receiver query $Q$,
\begin{equation}
    \operatorname{Attn}(Q,\widehat K,\widehat V)
    =
    \operatorname{softmax}\!\left(\frac{Q\widehat K^\top}{\sqrt{d_k}}\right)\widehat V.
\end{equation}
A perturbation to K changes where the query places attention; a perturbation to V changes what is retrieved under those weights. These effects are query-dependent and coupled. Mean squared cache error instead averages over layers, heads, tokens, and dimensions, without measuring the receiver computation induced by the translated state.
We measure that receiver-level computation on a matched Qwen3-14B$\rightarrow$32B diagnostic drawn from the same seven evaluation panels. For 50 examples per panel, we compare the native receiver cache, the Ridge base paired with the step-600 CacheLink checkpoint, and the final CacheLink cache. Appendix~\ref{app:protocol} gives the sampling and metric definitions.

\begin{figure*}[!t]
    \centering
    \includegraphics[width=\textwidth]{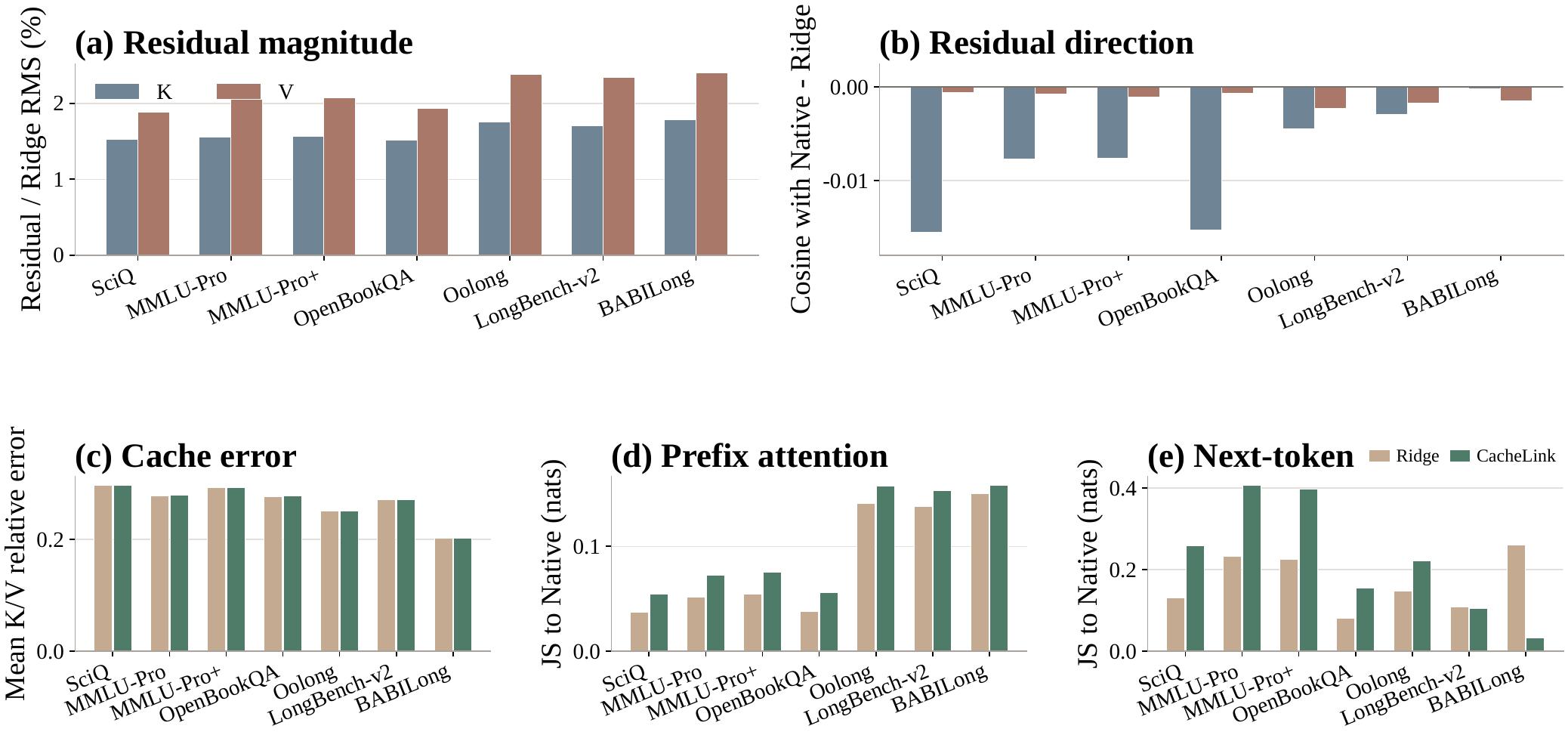}
    \caption{\textbf{A small residual changes receiver behavior while leaving reconstruction nearly unchanged.}
    The top row shows residual size and direction; the bottom row compares cache error, prefix attention, and next-token behavior.}
    \label{fig:fixed_panel_functional_decoupling}
\end{figure*}

\paragraph{A small residual changes receiver behavior while leaving reconstruction nearly unchanged.}
Figure~\ref{fig:fixed_panel_functional_decoupling} connects the geometry of the learned residual to the computation it induces in the receiver. Panels (a) and (b) show that the correction is small---only 1.52--1.79\% of Ridge RMS for K and 1.88--2.40\% for V---and is nearly orthogonal to the Native--Ridge gap across panels. Panel (c) then shows that adding this residual barely changes the average K/V reconstruction error. In contrast, panels (d) and (e) show clear changes in the receiver's prefix-attention and next-token distributions. The same small tensor-space correction can therefore have a much larger effect after it passes through receiver attention and prediction. On the separate 200-example short-prefix multiple-choice subset, length-normalized accuracy rises from 48\% to 51\%, with 12 wrong-to-correct and 6 correct-to-wrong flips.

\begin{figure*}[!t]
    \centering
    \includegraphics[width=0.98\textwidth]{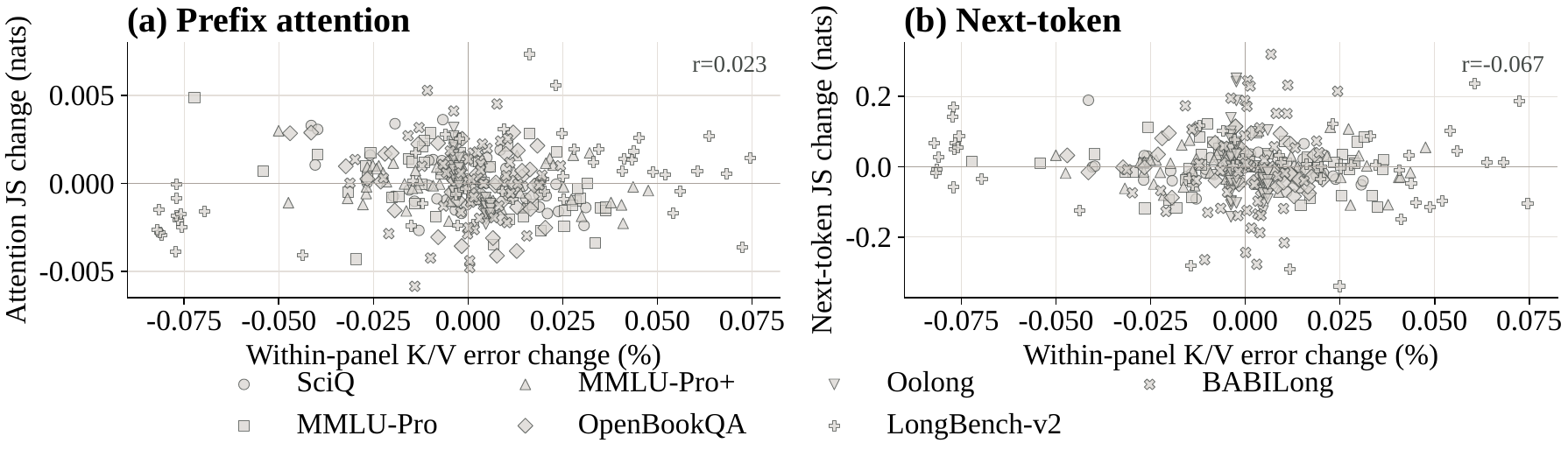}
    \caption{\textbf{Cache-error changes do not track receiver-level changes within panels.}
    Points are matched examples after within-panel centering.}
    \label{fig:fixed_panel_error_behavior}
\end{figure*}

\paragraph{The decoupling persists example by example.}
Figure~\ref{fig:fixed_panel_error_behavior} removes panel offsets by centering each metric within its panel before pooling examples. The centered Pearson association between mean K/V error reduction and attention-JS change is $-0.023$; the association with next-token-JS change is $0.067$. On the short-prefix panels, the association with normalized gold-answer log-probability change is $0.181$. Movement toward the native cache in squared-error space therefore carries little information about how receiver behavior changes.

\textbf{Observation 2.} Receiver-level functional change is largely decoupled from movement toward the native cache in squared-error space.

Observations 1 and 2 assign reconstruction a clear role: it provides useful structure, but receiver behavior must supply the optimization signal.

\FloatBarrier

\FloatBarrier

\subsection{K and V share a common latent origin}
\label{sec:kv_coupling}

At each layer and token, K and V are different projections of the same underlying hidden representation. Suppressing normalization and positional transforms for clarity, they take the form \(K=hW_K\) and \(V=hW_V\). Cross-model shifts in K and V can therefore share factors inherited from \(h\). Their functional roles are coupled as well: K determines the attention weights applied to V, so the effect of correcting one component depends on the state of the other.

\begin{figure*}[!t]
    \centering
    \includegraphics[width=0.98\textwidth]{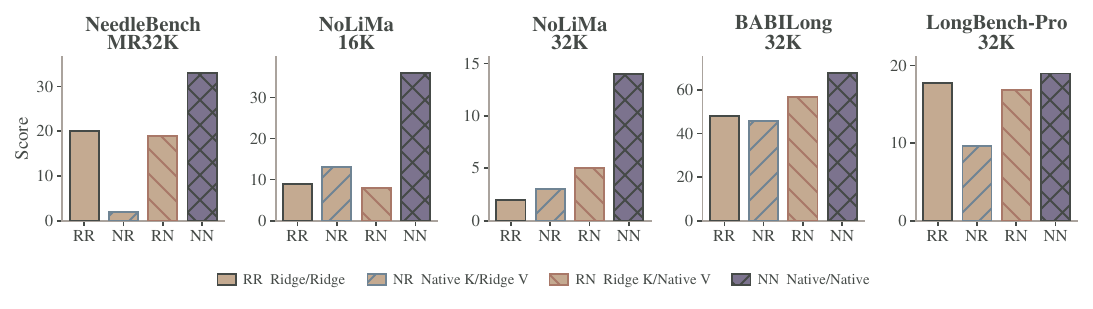}
    \caption{\textbf{K and V interact functionally under cross-model cache handoff.}
    RR, NR, RN, and NN denote Ridge K/Ridge V, Native K/Ridge V, Ridge K/Native V, and Native K/Native V, respectively. Replacing only one component does not produce a uniform improvement across tasks, so the effect of K depends on the paired V state and vice versa.}
    \label{fig:kv_source_swap}
\end{figure*}

We test this interaction with a targeted source-swap intervention on five tasks. Figure~\ref{fig:kv_source_swap} compares Ridge K/Ridge V (RR), Native K/Ridge V (NR), Ridge K/Native V (RN), and Native K/Native V (NN). RR is the deployable Ridge state; NR, RN, and NN are oracle interventions that expose the effect of replacing one or both cache components.
Replacing only K or only V does not yield a consistent improvement across tasks. Correcting one component can help or fail depending on the paired state of the other, matching their shared origin and coupled use in attention.

\textbf{Observation 3.} K and V transport errors share latent structure and should be corrected jointly.

Each observation maps directly to a CacheLink design choice. Ridge supplies the structural initialization; receiver-level supervision optimizes functional utility; and separate K/V stems followed by late fusion capture projection-specific and shared structure in a joint correction. Section~\ref{sec:method} implements these choices.

\FloatBarrier

%% file: sections/method.tex
\section{CacheLink: functional residual KV transport}
\label{sec:method}

The three observations in Section~\ref{sec:motivation} define the design of \method{}. Ridge initializes a complete receiver cache; separate K/V stems preserve projection-specific structure before late fusion forms a joint correction; and functional supervision trains the transported state through the frozen receiver with continuation CE and forward KL.

\Needspace{12\baselineskip}
\noindent
\begin{minipage}[t]{0.57\columnwidth}
\vspace{0pt}
\textbf{Ridge as structural initialization.}
A substantial fraction of receiver KV structure is linearly predictable from the sender cache. In the layerwise diagnostic summarized in Figure~\ref{fig:method_ridge_linearity}, mean held-out $R^2$ across the 64 receiver layers is 0.838 for K and 0.609 for V. Ridge therefore provides a complete, structured starting state; CacheLink uses its additional capacity to optimize the functional correction that reconstruction alone leaves unresolved.
\end{minipage}\hfill
\begin{minipage}[t]{0.40\columnwidth}
\vspace{0pt}
\centering
\includegraphics[width=\linewidth]{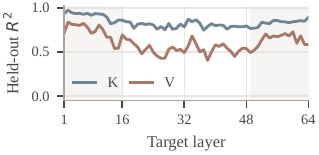}
\vspace{-0.50\baselineskip}
\captionof{figure}{\textbf{Linear KV predictability.}}
\label{fig:method_ridge_linearity}
\end{minipage}

\smallskip

For target layer $l$, let $\mathcal{S}_l$ contain the same $k=8$ sender layers selected by the Ridge baseline. At token position $t$, we form
\begin{equation}
  \begin{aligned}
  X^K_{l,t} &= \operatorname{concat}_{j\in\mathcal{S}_l}K^S_{j,t},
  \quad
  X^V_{l,t} &= \operatorname{concat}_{j\in\mathcal{S}_l}V^S_{j,t}.
  \end{aligned}
\end{equation}
where source keys are represented after kNorm and de-RoPE and values remain in raw cache space. The fitted Ridge maps produce the receiver-cache initialization,
\begin{equation}
  \begin{aligned}
  B^K_{l,t} &= R^K_l(X^K_{l,t}),
  \qquad
  B^V_{l,t} &= R^V_l(X^V_{l,t}).
  \end{aligned}
\end{equation}
\method{} predicts residual corrections $\Delta K$ and $\Delta V$ on top of this initialization. The residual output heads are zero-initialized, so every checkpoint begins from the Ridge solution. The fitted Ridge maps remain fixed throughout CacheLink training; only the residual mapper parameters are optimized.

\paragraph{Coupled K/V residual transport.}
K and V are produced from the same layer input hidden representation through different projections. We preserve their projection-specific structure with separate stems,
\begin{equation}
  z^K_{l,t}
  =
  \operatorname{GELU}(W^K_lX^K_{l,t}+b^K_l),
  \qquad
  z^V_{l,t}
  =
  \operatorname{GELU}(W^V_lX^V_{l,t}+b^V_l).
\end{equation}
The two views are then fused before either correction is produced:
\begin{align}
  u_{l,t}
  &=
  \operatorname{GELU}\!\left(W^F_l[z^K_{l,t};z^V_{l,t}]+b^F_l\right),\\
  \Delta K_{l,t}
  &=
  H^K_lu_{l,t},
  \qquad
  \Delta V_{l,t}
  =
  H^V_lu_{l,t}.
\end{align}
The shared representation \(u_{l,t}\) captures latent factors jointly expressed through K and V. The resulting receiver cache is
\begin{equation}
  \widehat K^T_{l,t}
  =
  \operatorname{RoPE}_T(B^K_{l,t}+\Delta K_{l,t}),
  \qquad
  \widehat V^T_{l,t}
  =
  B^V_{l,t}+\Delta V_{l,t}.
\end{equation}

\paragraph{Functional supervision through the receiver.}
We inject the translated cache into a frozen receiver and optimize the behavior induced by that state. For a continuation $y_{1:T}$ following the reusable prefix, the training objective is
\begin{equation}
  \mathcal{L}_{\mathrm{func}}
  =
  -\frac{1}{T}\sum_t \log p_{\mathrm{map}}(y_t)
  +
  \lambda_{\mathrm{KL}}
  \frac{1}{T}\sum_t
  D_{\mathrm{KL}}
  \!\left(
  p_{\mathrm{native},t}
  \,\Vert\,
  p_{\mathrm{map},t}
  \right).
\end{equation}
The CE term trains the transported state to support the target continuation; forward KL transfers the predictive behavior of a native receiver that prefills the same prefix. Gradients pass through receiver attention into \method{}, while receiver parameters remain frozen. The final objective contains no cache-reconstruction loss. Appendix~\ref{app:protocol} reports the loss ablation and implementation details.

\paragraph{Zero-prefill deployment.}
At deployment, the sender computes the reusable prefix once, \method{} translates the resulting cache once, and the receiver processes only the new query and generated tokens. The receiver never replays the transferred prefix.

%% file: sections/experiments.tex
\section{Experiments}
\label{sec:experiments}

\subsection{Experimental setup}
\label{sec:exp_setup}

\paragraph{Models and zero-prefill protocol.}
We evaluate four sender$\rightarrow$receiver pairs: Qwen3-14B$\rightarrow$32B and Qwen3-8B$\rightarrow$14B~\citep{yang2025qwen3}, Ministral-3-8B-Reasoning-2512$\rightarrow$Ministral-3-14B-Reasoning-2512~\citep{liu2026ministral3}, and Yi-1.5-9B$\rightarrow$34B~\citep{ai2024yi}. Every handoff method constructs a complete receiver cache from the sender cache and performs zero receiver prefill on the reusable prefix. Keys are translated after source de-RoPE and receive target RoPE before injection; values remain in raw cache space. Within each model pair and evaluation panel, all methods use identical tokenized inputs, prompts, decoding rules, and scorers.

\paragraph{Evaluation panels.}
We use seven fixed panels spanning short- and long-prefix settings. SciQ~\citep{welbl2017sciq}, MMLU-Pro~\citep{wang2024mmlupro}, MMLU-Pro-Plus~\citep{taghanaki2024mmluproplus}, and OpenBookQA~\citep{mihaylov2018openbookqa} are short-prefix multiple-choice tasks. Oolong-Synth 32K~\citep{bertsch2025oolong}, LongBench-v2~\citep{bai2024longbenchv2}, and BABILong 32K~\citep{kuratov2024babilong} use long reusable prefixes and task-specific scorers. MMLU-Pro and MMLU-Pro-Plus share question stems and are reported separately. The equal-weight mean provides a compact within-pair summary of the seven panel scores.

\paragraph{Baselines.}
We compare \method{} with the native sender, native receiver, Ridge-$k8$~\citep{heo2026crossmodel}, CacheBridge~\citep{qu2026cachebridge}, and LatentAlign~\citep{dery2026latent}. Native receiver re-prefills the reusable prefix and serves as the large-model reference; all cache-transport methods operate under the same zero-prefill protocol.

\input{data/four_model_pair_tables}
\FloatBarrier

\paragraph{Training.}
All four \method{} checkpoints use receiver continuation CE and forward KL as functional supervision. The Ridge maps are fitted before CacheLink training and remain frozen for every model pair; CacheLink optimizes only the residual mapper. Qwen3-14B$\rightarrow$32B trains its late K/V residual on 400 earlier samples plus 200 LongAlign reasoning samples~\citep{bai2024longalign}. Ministral-3-8B$\rightarrow$14B trains the six-source residual over its fixed Ridge cache. Qwen3-8B$\rightarrow$14B and Yi-1.5-9B$\rightarrow$34B train zero-initialized late K/V residuals for 600 optimizer steps on 1,200 examples, with two examples per step. Appendix~\ref{app:protocol} gives the complete training sources, cache conventions, and checkpoint dependencies.
The experiments test three claims: functional supervision recovers utility missed by reconstruction, the gain transfers across model pairs and prefix regimes, and zero-prefill transport retains its latency advantage.

\subsection{Functional supervision improves receiver utility}
\label{sec:main_results}

\method{} improves the equal-weight seven-panel mean over Ridge for every sender$\rightarrow$receiver pair: +6.84 points for Qwen3-14B$\rightarrow$32B, +2.96 for Ministral-3-8B$\rightarrow$14B, +16.01 for Qwen3-8B$\rightarrow$14B, and +11.14 for Yi-1.5-9B$\rightarrow$34B. At the panel level, it exceeds Ridge in 26 of 28 comparisons. The same aggregate pattern holds against the native sender, CacheBridge, and LatentAlign: \method{} has the higher equal-weight mean for all four model pairs and exceeds the native sender in 21 of 28 pair--panel comparisons.
These results match the mechanism in Section~\ref{sec:motivation}. Ridge supplies a strong reconstruction-based state, while receiver-level functional supervision recovers utility that reconstruction error does not identify.

\subsection{The gain transfers across model pairs and prefix regimes}
\label{sec:generality}

The improvement is not confined to one model family or context regime. On Qwen3-14B$\rightarrow$32B, Qwen3-8B$\rightarrow$14B, and Yi-1.5-9B$\rightarrow$34B, \method{} improves over Ridge on both the short-prefix and long-prefix panels. Ministral-3-8B$\rightarrow$14B shows the concentrated exception: \method{} improves all four short-prefix panels and BABILong, while Ridge remains stronger on Oolong-Synth and LongBench-v2.

The cross-pair pattern is therefore broad but task-dependent: functional transport improves both short- and long-context handoff across three model families, while the remaining regressions are concentrated in two long-prefix panels for one model pair.

\subsection{Zero-prefill transport preserves the latency advantage}
\label{sec:efficiency}

\begin{wrapfigure}{r}{0.60\columnwidth}
    \vspace{-0.7\baselineskip}
    \centering
    \includegraphics[width=\linewidth]{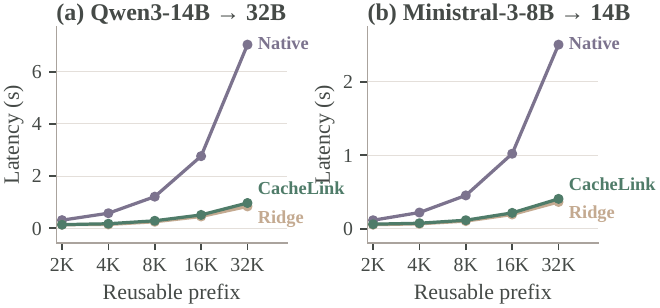}
    \caption{\textbf{Zero-prefill transport scales better with prefix length.}
    At 32K, CacheLink is $7.27\times$ and $6.18\times$ faster than receiver re-prefill on the two model pairs.}
    \label{fig:latency_scaling}
    \vspace{-0.8\baselineskip}
\end{wrapfigure}

We measure end-to-end handoff latency in the intended deployment setting: the sender has already processed the reusable prefix and its KV cache is available when a model switch occurs. Native receiver latency includes re-prefilling that prefix, whereas Ridge and \method{} translate the available sender cache before continuation. Figure~\ref{fig:latency_scaling} reports latency from 2K to 32K reusable prefixes on Qwen3-14B$\rightarrow$32B and Ministral-3-8B$\rightarrow$14B.

The residual correction adds modest overhead to Ridge while preserving most of the transport advantage. At 32K, latency increases from 0.839s for Ridge to 0.969s for \method{} on Qwen3-14B$\rightarrow$32B, and from 0.364s to 0.406s on Ministral-3-8B$\rightarrow$14B. Both remain far below receiver re-prefill, reaching $7.27\times$ and $6.18\times$ speedups respectively. The advantage grows with prefix length because re-prefill scales with the transferred context while cache translation avoids receiver-prefix replay.

%% file: data/four_model_pair_tables.tex
\begin{table*}[!t]
\caption{Main results across four sender$\rightarrow$receiver model pairs. Scores are percentages. CB and LA denote CacheBridge and LatentAlign, respectively. Native receiver is shown as the large-model reference and is excluded from boldface selection. Bold marks the best among native sender, Ridge-$k8$, CB, LA, and \method{}. Equal-weight means summarize the seven panel scores and are not pooled accuracy.}
\label{tab:four_pair_main}
\centering
\scriptsize
\setlength{\tabcolsep}{5pt}
\renewcommand{\arraystretch}{1.04}
\resizebox{\textwidth}{!}{%
\begin{tabular}{@{}lr@{\hspace{5pt}\vrule\hspace{5pt}}rrrrr@{}}
\toprule
Dataset / panel & Native receiver & Native sender & Ridge-$k8$ & CB & LA & \method{} \\
\midrule
\rowcolor{gray!12}\multicolumn{7}{@{}l}{\textbf{Qwen3 14B$\rightarrow$32B}} \\
SciQ & 73.30 & 59.70 & 74.80 & 60.20 & 25.50 & \textbf{83.90} \\
MMLU-Pro & 48.96 & 36.36 & 41.20 & 36.54 & 11.10 & \textbf{47.46} \\
MMLU-Pro-Plus & 36.03 & 25.08 & 35.75 & 29.39 & 6.40 & \textbf{40.32} \\
OpenBookQA & 39.20 & 37.80 & 40.20 & 39.00 & 28.00 & \textbf{42.20} \\
Oolong-Synth 32K & 29.00 & 14.23 & 26.55 & 25.75 & 0.00 & \textbf{38.12} \\
LongBench-v2 & 48.28 & 43.10 & 42.24 & 35.34 & 0.00 & \textbf{44.83} \\
BABILong 32K & 62.20 & 54.60 & 43.60 & 0.00 & 0.00 & \textbf{55.40} \\
\emph{Equal-weight mean} & 48.14 & 38.70 & 43.48 & 32.32 & 10.14 & \textbf{50.32} \\
\midrule
\rowcolor{gray!12}\multicolumn{7}{@{}l}{\textbf{Ministral 3 8B$\rightarrow$14B}} \\
SciQ & 54.30 & 65.60 & 59.10 & 50.90 & 22.40 & \textbf{72.60} \\
MMLU-Pro & 40.46 & 28.10 & 30.49 & 21.89 & 11.69 & \textbf{34.97} \\
MMLU-Pro-Plus & 46.98 & 22.17 & 19.71 & 12.11 & 6.57 & \textbf{23.45} \\
OpenBookQA & 35.60 & 36.40 & 35.20 & 33.60 & 26.60 & \textbf{38.80} \\
Oolong-Synth 32K & 42.86 & 38.20 & \textbf{42.01} & 33.81 & 0.00 & 36.14 \\
LongBench-v2 & 45.69 & 37.93 & \textbf{42.24} & 38.79 & 0.00 & 39.66 \\
BABILong 32K & 50.80 & \textbf{51.60} & 41.20 & 37.80 & 0.00 & 45.00 \\
\emph{Equal-weight mean} & 45.24 & 40.00 & 38.56 & 32.70 & 9.61 & \textbf{41.52} \\
\midrule
\rowcolor{gray!12}\multicolumn{7}{@{}l}{\textbf{Qwen3 8B$\rightarrow$14B}} \\
SciQ & 59.70 & 48.10 & 47.50 & 48.70 & 21.80 & \textbf{81.90} \\
MMLU-Pro & 36.34 & 40.49 & 20.78 & 11.81 & 11.35 & \textbf{44.28} \\
MMLU-Pro-Plus & 25.08 & \textbf{33.50} & 15.51 & 6.61 & 6.32 & 30.00 \\
OpenBookQA & 37.80 & 36.60 & 34.40 & 31.20 & 26.40 & \textbf{39.80} \\
Oolong-Synth 32K & 47.78 & 41.11 & 41.20 & 3.72 & 0.00 & \textbf{43.02} \\
LongBench-v2 & 39.66 & \textbf{38.79} & 31.90 & 32.76 & 0.00 & 37.93 \\
BABILong 32K & 53.80 & \textbf{51.00} & 9.00 & 0.00 & 0.00 & 35.40 \\
\emph{Equal-weight mean} & 42.88 & 41.37 & 28.61 & 19.26 & 9.41 & \textbf{44.62} \\
\midrule
\rowcolor{gray!12}\multicolumn{7}{@{}l}{\textbf{Yi-1.5 9B$\rightarrow$34B}} \\
SciQ & 80.00 & 72.70 & 62.00 & 51.80 & 23.90 & \textbf{81.00} \\
MMLU-Pro & 41.40 & 31.57 & 24.35 & 11.87 & 11.52 & \textbf{34.18} \\
MMLU-Pro-Plus & 41.57 & \textbf{35.25} & 18.09 & 6.70 & 6.51 & 24.69 \\
OpenBookQA & 44.20 & 38.40 & 33.60 & 31.60 & 28.80 & \textbf{41.20} \\
Oolong-Synth 32K & 20.26 & 1.61 & 12.84 & 3.75 & 0.00 & \textbf{26.67} \\
LongBench-v2 & 31.90 & 23.28 & 22.41 & 7.76 & 0.00 & \textbf{30.17} \\
BABILong 32K & 41.40 & \textbf{37.40} & 8.40 & 3.40 & 0.00 & 21.80 \\
\emph{Equal-weight mean} & 42.96 & 34.32 & 25.96 & 16.70 & 10.10 & \textbf{37.10} \\
\bottomrule
\end{tabular}
}
\end{table*}

%% file: sections/conclusion.tex
\section{Conclusion}

Cross-model KV states can be substantially reconstructed, but reconstruction fidelity does not determine whether the receiver can use the transported state effectively. \method{} turns this observation into a transport objective: it starts from a Ridge-initialized receiver cache, couples K/V corrections through late fusion, and trains the transported state through the frozen receiver. Across four sender$\rightarrow$receiver pairs and seven fixed evaluation panels, \method{} improves the equal-weight mean over both Ridge and the native sender for every pair, exceeding Ridge in 26 of 28 pair--panel comparisons and the sender in 21 of 28. Matched diagnostics show why: CacheLink changes K/V squared error by less than 0.4\% while substantially changing receiver attention and next-token behavior. Zero-prefill cross-model handoff is therefore better framed as preserving receiver computation than as reconstructing receiver tensors.

%% file: sections/appendix.tex
\section{Evaluation and diagnostic protocol}
\label{app:protocol}

\paragraph{Cache transport.}
Qwen3-14B is the sender and Qwen3-32B is the receiver for the mechanism diagnostic. The receiver has 64 transformer layers; both models expose eight KV heads of dimension 128. A handoff constructs a complete receiver cache and does not prefill the receiver on the reusable prefix. Keys are mapped in de-RoPE content space and receive the receiver's RoPE after transport; values are mapped in raw cache space.

\paragraph{CacheLink architecture.}
For Qwen3-14B$\rightarrow$32B, concatenating the eight selected source layers gives 8,192-dimensional K and V views. The two stems are $8192\!\rightarrow\!256$, the late-fusion layer is $512\!\rightarrow\!512$, and two $512\!\rightarrow\!1024$ heads emit all eight target K or V heads. The residual mapper has 5,508,096 trainable parameters per target layer and 352,518,144 across the 64-layer receiver. It operates directly on the sender cache and requires neither a source-side solo forward pass nor receiver replay of the transferred prefix.

\paragraph{Ridge fits and step-600 checkpoints.}
The main-table Ridge baseline is calibrated on 500 FineWeb-Edu sequences of 1,024 tokens, with $k=8$ selected source layers, all eight source KV heads, and a separate affine regressor for each target layer/head and for K/V. For every model pair, these Ridge maps are fitted before CacheLink training and then frozen. The Qwen3-14B$\rightarrow$32B step-600 CacheLink checkpoint trains only the late K/V residual on 400 earlier samples plus 200 LongAlign reasoning samples using receiver continuation cross-entropy and forward KL; inference loads the fixed \texttt{ridge/} fit together with the residual. The Ministral-3-8B$\rightarrow$14B checkpoint trains the original six-source residual over its fixed Ridge cache with receiver cross-entropy plus forward KL. The Qwen3-8B$\rightarrow$14B and Yi-1.5-9B$\rightarrow$34B checkpoints likewise train zero-initialized late K/V residuals over fixed pair-specific Ridge caches with receiver cross-entropy plus forward KL. Each of the latter two runs uses 1,200 training examples, two per optimizer step for 600 steps, drawn from CommonsenseQA, CosmosQA, SQuAD v2, DROP, MultiDoc2Dial, and ContractNLI. These are training sources, not additional evaluation panels.

\paragraph{Why CE + forward KL?}
We selected receiver continuation cross-entropy plus forward KL as the shared functional objective based on an earlier controlled loss ablation for Qwen3-14B$\rightarrow$32B. This pilot used the earlier CacheLink architecture for 150 optimizer steps while holding the training data, fixed Ridge fit, model structure, and optimization settings constant. All variants were evaluated on the same 100-question NeedleBench MR32K set with a 4,096-token output cap and strict boxed-answer matching. Table~\ref{tab:loss_ablation} reports the archived results.

\begin{table}[!t]
\centering
\small
\begin{tabular}{lc}
\toprule
Training objective & Correct / 100 \\
\midrule
CE only & 37 \\
CE + forward KL & \textbf{40} \\
CE + reverse KL & 39 \\
Forward KL only & 29 \\
Reverse KL only & 23 \\
\midrule
Ridge & 28 \\
Native sender (14B) & 36 \\
Native receiver (32B) & 59 \\
\bottomrule
\end{tabular}
\caption{\textbf{Loss-function ablation on Qwen3-14B$\rightarrow$32B.}
}
\label{tab:loss_ablation}
\end{table}

The ablation separates the roles of the two functional terms. CE alone already improves over Ridge and the native sender, showing that continuation supervision directly teaches the transported state to support the target task. KL alone is much weaker, but adding forward KL to CE improves the score from 37 to 40, suggesting that matching the native receiver distribution supplies useful supervision beyond the gold continuation. Forward KL also performs slightly better than reverse KL in the matched CE setting (40 versus 39). KL without CE performs substantially worse. We therefore use CE + forward KL as the functional objective in the main experiments for all four model pairs.

\paragraph{Fixed evaluation panels.}
All reported evaluation results use the seven panels in Table~\ref{tab:four_pair_main}. SciQ, MMLU-Pro, MMLU-Pro-Plus, and OpenBookQA use their complete selected test splits. Oolong-Synth uses the fixed 100-question 32K panel for Qwen3-14B$\rightarrow$32B and Ministral-3-8B$\rightarrow$14B, and a fixed 400-question panel for Qwen3-8B$\rightarrow$14B and Yi-1.5-9B$\rightarrow$34B. LongBench-v2 uses the pair-specific frozen 8K--32K tokenized panel; BABILong uses the 500-example 32K QA1--QA5 panel. Exact tokenized-panel hashes and completion records are retained with the evaluation artifacts. MMLU-Pro and MMLU-Pro-Plus share question stems. Different task metrics, panel sizes, tokenizers, and preprocessing mean cross-panel means are descriptive summaries, not pooled accuracy.

\begin{figure*}[!t]
    \centering
    \includegraphics[width=\textwidth]{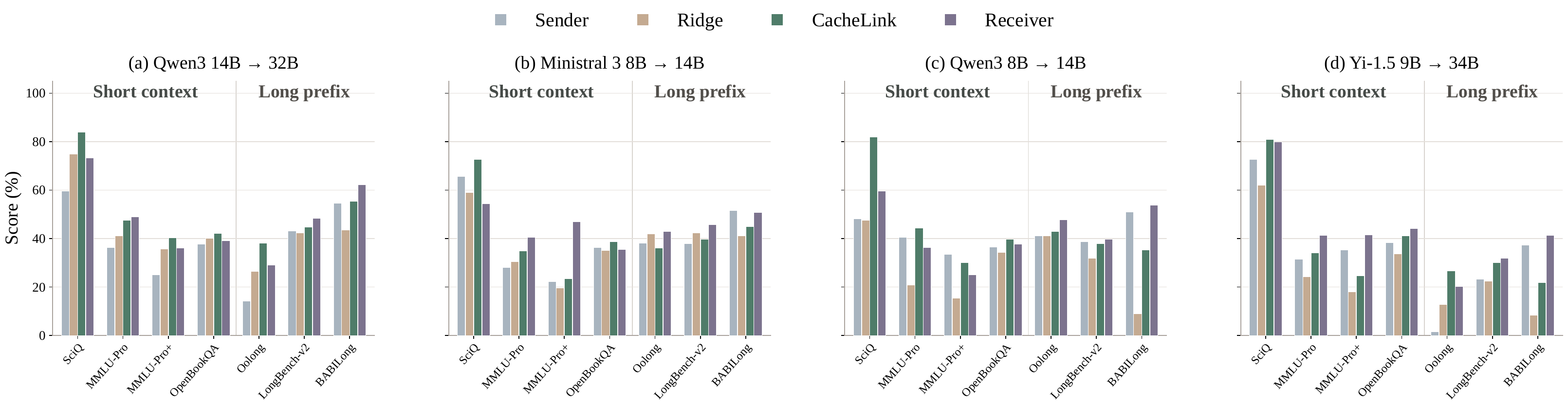}
    \caption{\textbf{Receiver utility under cross-model handoff is strongly task- and pair-dependent.}
    Each subplot shows the native sender, Ridge-$k8$, CacheLink, and native receiver on the same seven fixed panels. SciQ, MMLU-Pro, MMLU-Pro-Plus, and OpenBookQA are short-prefix multiple-choice panels; Oolong-Synth 32K, LongBench-v2, and BABILong 32K use long reusable prefixes.}
    \label{fig:ridge_short_long_four_pairs}
\end{figure*}

\paragraph{Cross-pair Ridge utility.}
Figure~\ref{fig:ridge_short_long_four_pairs} reports the complete seven-panel comparison for all four sender--receiver pairs. Ridge is competitive with the native receiver on some panels and falls below the native sender on others despite using the same mapping family throughout. Averaged across the four short-prefix and three long-prefix panels, the receiver-minus-Ridge gaps are 1.38 and 9.03 points for Qwen3-14B$\rightarrow$32B, 8.21 and 4.63 for Ministral-3-8B$\rightarrow$14B, 10.18 and 19.71 for Qwen3-8B$\rightarrow$14B, and 17.28 and 16.64 for Yi-1.5-9B$\rightarrow$34B. The variation is therefore pair- and task-dependent rather than a uniform long-context failure.

\paragraph{In-house baseline scope.}
We reproduce CacheBridge~\citep{qu2026cachebridge} and LatentAlign~\citep{dery2026latent} as cross-model KV transport baselines under the same zero-prefill evaluation protocol used for Ridge and \method{}. Table~\ref{tab:four_pair_main} reports their matched results on the same model pairs, inputs, prompts, decoding rules, and scorers.

\paragraph{Execution.}
Custom per-layer cache injection is not supported by the standard vLLM generation interface used in our environment. We therefore evaluate with Transformers and FlashAttention-2. The 14B sender and 32B receiver reside on separate H100 GPUs; a one-slot pipeline overlaps preparation of the next sender cache with receiver work on the current item. This changes throughput, not the tokenized inputs, decoding configuration, or scorer.

\paragraph{Matched mechanism diagnostic.}
For the Qwen3-14B$\rightarrow$32B pair, we select 50 examples from each of the seven fixed evaluation panels with seed 20260926, stratifying by category or task where metadata permits. This is a diagnostic subset and does not replace the complete main-table scores. On each example we compare the native receiver cache, the fixed Ridge cache used to initialize CacheLink, and the final cache after applying the step-600 residual.

We compute post-RoPE K and raw-V mean squared error, relative squared $L_2$ error, and global cosine similarity against the native cache. For the learned correction $\Delta=C_{\mathrm{CacheLink}}-C_{\mathrm{Ridge}}$, we report its RMS relative to the Ridge cache, its cosine with $C_{\mathrm{Native}}-C_{\mathrm{Ridge}}$, and the fraction of K/V squared error removed. We also compare final-query-token attention over reusable-prefix positions after renormalizing those probabilities within the prefix, using Jensen--Shannon divergence, entropy, and top-10 attention mass. This conditional distribution does not measure the total fraction of attention allocated to the prefix versus query tokens. The next-token distributions are compared by forward KL, Jensen--Shannon divergence, and top-1 agreement. These are token- and cache-level diagnostics: they do not measure full-generation accuracy. Multiple-choice answer scores are additionally computed for the four short-prefix panels using both raw and length-normalized continuation likelihood. 

\section{Supporting diagnostic analyses}
\label{app:legacy_diagnostics}

The following analyses provide additional evidence for three findings developed in the main text: Ridge already captures substantial linear sender--receiver KV structure; tensor reconstruction quality can remain nearly unchanged while task utility changes sharply; and a functionally useful residual need not behave like another step toward the native receiver cache. These results are used as targeted diagnostics rather than aggregate benchmark results.

\paragraph{Ridge provides a strong reconstruction base.}
The NeedleBench MR32K layerwise diagnostic measures how much receiver KV structure can be predicted linearly from the sender cache before any functional residual is introduced. Across the 64 receiver layers, mean held-out $R^2$ is 0.838 for K and 0.609 for V, with mean NMSE 0.108 and 0.364, respectively. The important point is not that Ridge reconstructs every layer perfectly, but that a large fraction of receiver-cache structure is already captured by a simple affine map. This observation motivates CacheLink's residual design: preserve the strong Ridge reconstruction and learn only what receiver behavior still requires.

\begin{figure*}[!t]
    \centering
    \includegraphics[width=0.98\textwidth]{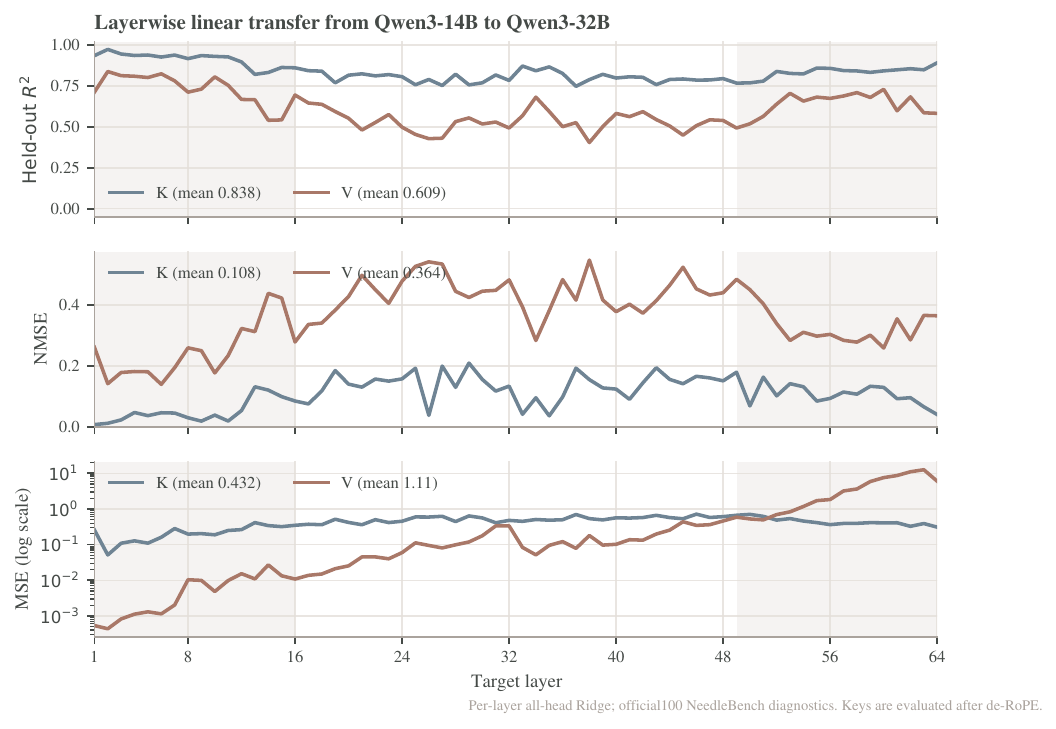}
    \caption{\textbf{Layerwise cache predictability for Qwen3-14B$\rightarrow$32B on NeedleBench MR32K.} The plot reports held-out K/V reconstruction statistics by receiver layer.}
    \label{fig:legacy_layerwise_linearity}
\end{figure*}

\paragraph{Nearly identical tensor error can correspond to very different task utility.}
The second pilot compares nine archived mapper variants while holding the evaluation task fixed. On NeedleBench MR32K, their KV-NMSE values occupy only 0.177239--0.177375, yet accuracy ranges from 12\% to 26\%. On NoLiMa 8K, KV-NMSE occupies 0.184919--0.185070 while accuracy ranges from 19\% to 32\%. Thus a mapper can move substantially in task quality without a correspondingly visible change in aggregate tensor error. This pilot supplied the first direct evidence for the reconstruction--utility mismatch; the matched seven-panel diagnostics in Section~\ref{sec:mse_mismatch} test the same idea more systematically.

\begin{figure*}[!t]
    \centering
    \includegraphics[width=0.98\textwidth]{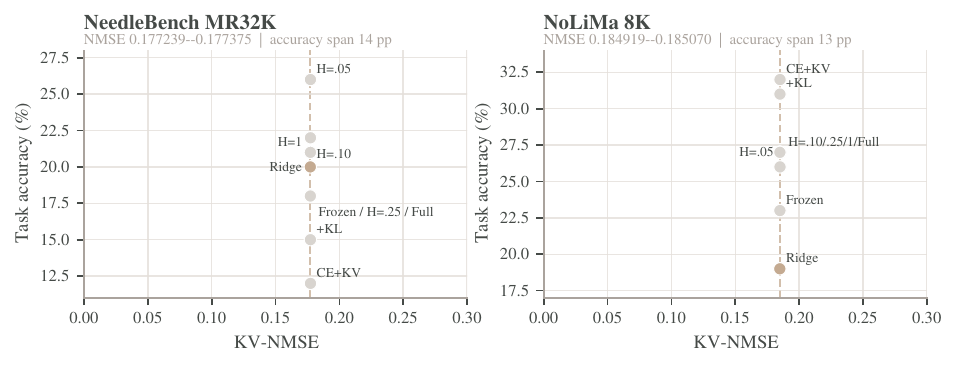}
    \caption{\textbf{Mapper-variant comparison on NeedleBench MR32K and NoLiMa 8K.} Similar aggregate KV-NMSE can accompany substantially different task scores.}
    \label{fig:legacy_nmse_utility}
\end{figure*}

\paragraph{Functional residuals need not be reconstruction residuals.}
The third pilot asks whether receiver-level training merely takes an additional step from Ridge toward the native receiver tensors. Its four panels measure the learned correction relative to the Ridge cache, relative to the Native-minus-Ridge gap, its cosine with the Native-minus-Ridge direction, and the post-/pre-correction Euclidean error ratio. If functional training were simply improving reconstruction, the correction would be expected to align with Native-minus-Ridge and systematically reduce the final tensor error. Instead, the layerwise analysis shows that useful corrections can remain small-to-moderate while not consistently following that reconstruction direction. This observation motivated the cleaner fixed-panel residual-direction analysis in Figure~\ref{fig:fixed_panel_functional_decoupling}.

\begin{figure*}[!t]
    \centering
    \includegraphics[width=0.98\textwidth]{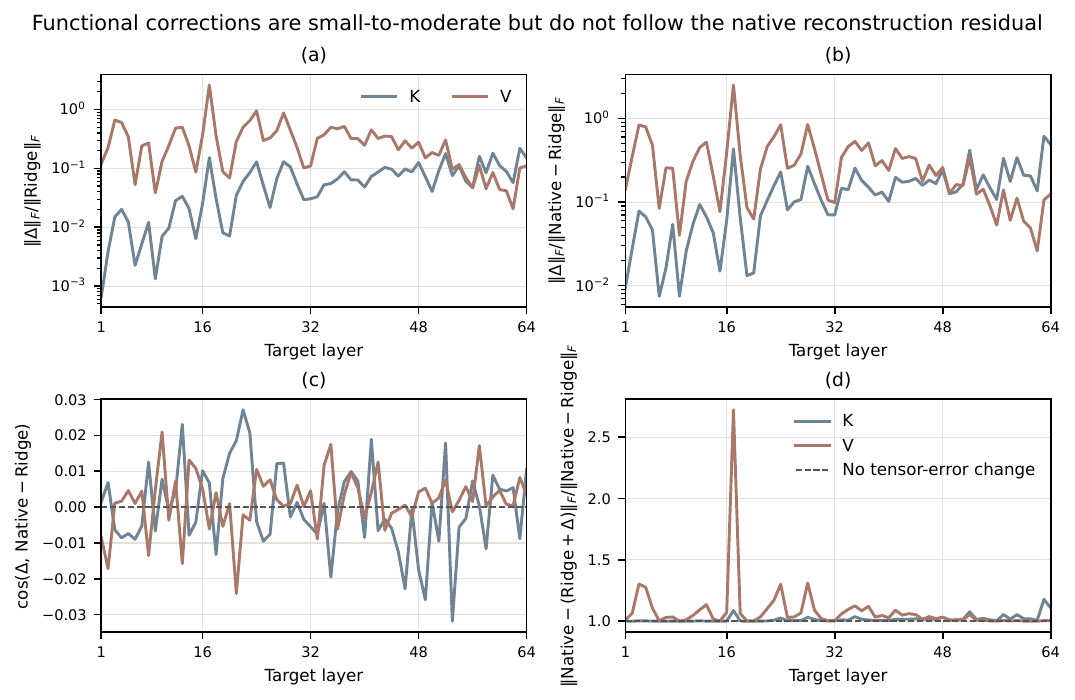}
    \caption{\textbf{Layerwise residual analysis on four held-out 32K LongABC documents.} The panels show correction magnitude, alignment with the Native-minus-Ridge direction, and the change in Euclidean error.}
    \label{fig:legacy_functional_residual}
\end{figure*}

%% file: references.bib
@article{heo2026crossmodel,
  title={Cross-Model KV Cache Transfer in LLM Families: A Closed-Form Linear Mapping for Prefill Reuse},
  author={Heo, Taekyung and Shafipour, Rasoul and Zhao, Ritchie and Golub, Maximilian and Kamani, Mohammad Mahdi and Borkar, Ritika and Chandran, Makesh Tarun and Zardoshti, Pantea and Rouhani, Bita Darvish},
  journal={arXiv preprint arXiv:2608.03893},
  year={2026}
}

@article{qu2026cachebridge,
  title={CacheBridge: Efficient Cross-Model KV Cache Transfer},
  author={Qu, Xingyu and Lu, Siyuan and Chen, Zhiyu and Wang, Sheng and Lin, Tao},
  journal={arXiv preprint arXiv:2609.00891},
  year={2026}
}

@inproceedings{bai2024longbenchv2,
  title={LongBench v2: Towards Deeper Understanding and Reasoning on Realistic Long-context Multitasks},
  author={Bai, Yushi and Tu, Shangqing and Zhang, Jiajie and Peng, Hao and Wang, Xiaozhi and Lv, Xin and Cao, Shulin and Xu, Jiazheng and Hou, Lei and Dong, Yuxiao and Tang, Jie and Li, Juanzi},
  booktitle={Proceedings of the 63rd Annual Meeting of the Association for Computational Linguistics},
  pages={3639--3664},
  year={2025}
}

@article{yang2025qwen3,
  title={Qwen3 Technical Report},
  author={Yang, An and Li, Anfeng and Yang, Baosong and Zhang, Beichen and Hui, Binyuan and Zheng, Bo and Yu, Bowen and Gao, Chang and others},
  journal={arXiv preprint arXiv:2505.09388},
  year={2025}
}

@article{kuratov2024babilong,
  title={BABILong: Testing the Limits of LLMs with Long Context Reasoning-in-a-Haystack},
  author={Kuratov, Yuri and Bulatov, Aydar and Anokhin, Petr and Rodkin, Ivan and Sorokin, Dmitry and Sorokin, Artyom and Burtsev, Mikhail},
  journal={arXiv preprint arXiv:2406.10149},
  year={2024}
}

@article{zhang2026evoroute,
  title={EvoRoute: Experience-Driven Self-Routing LLM Agent Systems},
  author={Zhang, Guibin and Yu, Haiyang and Yang, Kaiming and Wu, Bingli and Huang, Fei and Li, Yongbin and Yan, Shuicheng},
  journal={arXiv preprint arXiv:2601.02695},
  year={2026}
}

@inproceedings{fu2026c2c,
  title={Cache-to-Cache: Direct Semantic Communication Between Large Language Models},
  author={Fu, Tianyu and Min, Zihan and Zhang, Hanling and Yan, Jichao and Dai, Guohao and Ouyang, Wanli and Wang, Yu},
  booktitle={International Conference on Learning Representations},
  year={2026}
}

@article{dery2026latent,
  title={Latent Space Communication via K-V Cache Alignment},
  author={Dery, Lucio M. and Yahav, Zohar and Prior, Henry and Feng, Qixuan and Shen, Jiajun and Szlam, Arthur},
  journal={arXiv preprint arXiv:2601.06123},
  year={2026}
}

@inproceedings{zhao2025iam,
  title={IAM: Efficient Inference through Attention Mapping between Different-scale LLMs},
  author={Zhao, Yi and Li, Zuchao and Zhao, Hai},
  booktitle={Proceedings of the 63rd Annual Meeting of the Association for Computational Linguistics},
  pages={19522--19533},
  year={2025}
}

@inproceedings{liu2026droidspeak,
  title={DroidSpeak: KV Cache Sharing Across Fine-tuned Model Variants},
  author={Liu, Yuhan and Huang, Yuyang and Yao, Jiayi and Feng, Shaoting and Gu, Zhuohan and Du, Kuntai and Li, Hanchen and Cheng, Yihua and Jiang, Junchen and Lu, Shan and Musuvathi, Madan and Choukse, Esha},
  booktitle={23rd USENIX Symposium on Networked Systems Design and Implementation},
  pages={319--338},
  year={2026}
}

@inproceedings{woo2026icarus,
  title={ICaRus: Identical Cache Reuse for Efficient Multi-Model Inference},
  author={Woo, Sunghyeon and Kil, Jaeeun and Kim, Hoseung and Kim, Minsub and Kim, Joonghoon and Seo, Ahreum and Lee, Sungjae and Jo, Minjung and Ryu, Jiwon and Park, Baeseong and Kwon, Se Jung and Lee, Dongsoo},
  booktitle={International Conference on Learning Representations},
  year={2026}
}

@article{woo2026prefillshare,
  title={PrefillShare: A Shared Prefill Module for KV Reuse in Multi-LLM Disaggregated Serving},
  author={Woo, Sunghyeon and Kim, Hoseung and Shim, Sunghwan and Jo, Minjung and Jeong, Hyunjoon and Lee, Jeongtae and Kim, Joonghoon and Lee, Sungjae and Park, Baeseong and Kwon, Se Jung and Lee, Dongsoo},
  journal={arXiv preprint arXiv:2602.12029},
  year={2026}
}

@inproceedings{gim2024promptcache,
  title={Prompt Cache: Modular Attention Reuse for Low-Latency Inference},
  author={Gim, In and Chen, Guojun and Lee, Seung-seob and Sarda, Nikhil and Khandelwal, Anurag and Zhong, Lin},
  booktitle={Proceedings of Machine Learning and Systems},
  volume={6},
  year={2024}
}

@inproceedings{yao2025cacheblend,
  title={CacheBlend: Fast Large Language Model Serving for RAG with Cached Knowledge Fusion},
  author={Yao, Jiayi and Li, Hanchen and Liu, Yuhan and Ray, Siddhant and Cheng, Yihua and Zhang, Qizheng and Du, Kuntai and Lu, Shan and Jiang, Junchen},
  booktitle={Proceedings of the Twentieth European Conference on Computer Systems},
  year={2025}
}

@inproceedings{srivatsa2025preble,
  title={Preble: Efficient Distributed Prompt Scheduling for LLM Serving},
  author={Srivatsa, Vikranth and He, Zijian and Abhyankar, Reyna and Li, Dongming and Zhang, Yiying},
  booktitle={International Conference on Learning Representations},
  year={2025}
}

@inproceedings{zheng2024sglang,
  title={SGLang: Efficient Execution of Structured Language Model Programs},
  author={Zheng, Lianmin and Yin, Liangsheng and Xie, Zhiqiang and Sun, Chuyue and Huang, Jeff and Yu, Cody Hao and Cao, Shiyi and Kozyrakis, Christos and Stoica, Ion and Gonzalez, Joseph E. and Barrett, Clark and Sheng, Ying},
  booktitle={Advances in Neural Information Processing Systems},
  volume={37},
  year={2024}
}

@inproceedings{qin2025mooncake,
  title={Mooncake: Trading More Storage for Less Computation -- A KVCache-centric Architecture for Serving LLM Chatbot},
  author={Qin, Ruoyu and Li, Zheming and He, Weiran and Cui, Jialei and Ren, Feng and Zhang, Mingxing and Wu, Yongwei and Zheng, Weimin and Xu, Xinran},
  booktitle={23rd USENIX Conference on File and Storage Technologies},
  pages={155--170},
  year={2025}
}

@inproceedings{kwon2023vllm,
  title={Efficient Memory Management for Large Language Model Serving with PagedAttention},
  author={Kwon, Woosuk and Li, Zhuohan and Zhuang, Siyuan and Sheng, Ying and Zheng, Lianmin and Yu, Cody Hao and Gonzalez, Joseph E. and Zhang, Hao and Stoica, Ion},
  booktitle={Proceedings of the 29th Symposium on Operating Systems Principles},
  pages={611--626},
  year={2023}
}

@inproceedings{liu2024kivi,
  title={KIVI: A Tuning-Free Asymmetric 2bit Quantization for KV Cache},
  author={Liu, Zirui and Yuan, Jiayi and Jin, Hongye and Zhong, Shaochen and Xu, Zhaozhuo and Braverman, Vladimir and Chen, Beidi and Hu, Xia},
  booktitle={Proceedings of the 41st International Conference on Machine Learning},
  pages={32332--32344},
  year={2024}
}

@inproceedings{tang2024quest,
  title={QUEST: Query-Aware Sparsity for Efficient Long-Context LLM Inference},
  author={Tang, Jiaming and Zhao, Yilong and Zhu, Kan and Xiao, Guangxuan and Kasikci, Baris and Han, Song},
  booktitle={Proceedings of the 41st International Conference on Machine Learning},
  pages={47901--47911},
  year={2024}
}

@article{li2024snapkv,
  title={SnapKV: LLM Knows What You are Looking for Before Generation},
  author={Li, Yuhong and Huang, Yingbing and Yang, Bowen and Venkitesh, Bharat and Locatelli, Acyr and Ye, Hanchen and Cai, Tianle and Lewis, Patrick and Chen, Deming},
  journal={arXiv preprint arXiv:2404.14469},
  year={2024}
}

@inproceedings{ong2025routellm,
  title={RouteLLM: Learning to Route LLMs with Preference Data},
  author={Ong, Isaac and Almahairi, Amjad and Wu, Vincent and Chiang, Wei-Lin and Wu, Tianhao and Gonzalez, Joseph E. and Kadous, M. Waleed and Stoica, Ion},
  booktitle={International Conference on Learning Representations},
  year={2025}
}

@article{chen2024frugalgpt,
  title={FrugalGPT: How to Use Large Language Models While Reducing Cost and Improving Performance},
  author={Chen, Lingjiao and Zaharia, Matei and Zou, James},
  journal={Transactions on Machine Learning Research},
  year={2024}
}

@inproceedings{aggarwal2024automix,
  title={AutoMix: Automatically Mixing Language Models},
  author={Aggarwal, Pranjal and Madaan, Aman and Anand, Ankit and Potharaju, Srividya Pranavi and Mishra, Swaroop and Zhou, Pei and Gupta, Aditya and Rajagopal, Dheeraj and Kappaganthu, Karthik and Yang, Yiming and Upadhyay, Shyam and Faruqui, Manaal and Mausam},
  booktitle={Advances in Neural Information Processing Systems},
  volume={37},
  year={2024}
}

@inproceedings{hu2024routerbench,
  title={RouterBench: A Benchmark for Multi-LLM Routing System},
  author={Hu, Qitian Jason and Bieker, Jacob and Li, Xiuyu and Jiang, Nan and Keigwin, Benjamin and Ranganath, Gaurav and Keutzer, Kurt and Upadhyay, Shriyash Kaustubh},
  booktitle={ICML 2024 Workshops: Agentic Markets},
  year={2024}
}

@article{dohan2022cascades,
  title={Language Model Cascades},
  author={Dohan, David and Xu, Winnie and Lewkowycz, Aitor and Austin, Jacob and Bieber, David and Lopes, Raphael Gontijo and Wu, Yuhuai and Michalewski, Henryk and Saurous, Rif A. and Sohl-Dickstein, Jascha and Murphy, Kevin and Sutton, Charles},
  journal={arXiv preprint arXiv:2207.10342},
  year={2022}
}

@inproceedings{chen2025stitching,
  title={Transferring Linear Features Across Language Models With Model Stitching},
  author={Chen, Alan and Merullo, Jack and Stolfo, Alessandro and Pavlick, Ellie},
  booktitle={Advances in Neural Information Processing Systems},
  volume={38},
  year={2025}
}

@inproceedings{bansal2021stitching,
  title={Revisiting Model Stitching to Compare Neural Representations},
  author={Bansal, Yamini and Nakkiran, Preetum and Barak, Boaz},
  booktitle={Advances in Neural Information Processing Systems},
  volume={34},
  year={2021}
}

@inproceedings{kornblith2019cka,
  title={Similarity of Neural Network Representations Revisited},
  author={Kornblith, Simon and Norouzi, Mohammad and Lee, Honglak and Hinton, Geoffrey},
  booktitle={Proceedings of the 36th International Conference on Machine Learning},
  pages={3519--3529},
  year={2019}
}

@inproceedings{huh2024platonic,
  title={Position: The Platonic Representation Hypothesis},
  author={Huh, Minyoung and Cheung, Brian and Wang, Tongzhou and Isola, Phillip},
  booktitle={Proceedings of the 41st International Conference on Machine Learning},
  pages={20617--20642},
  year={2024}
}

@article{liu2026ministral3,
  title={Ministral 3},
  author={Liu, Alexander H. and Khandelwal, Kartik and Subramanian, Sandeep and Jouault, Victor and Rastogi, Abhinav and others},
  journal={arXiv preprint arXiv:2601.08584},
  year={2026}
}

@article{ai2024yi,
  title={Yi: Open Foundation Models by 01.AI},
  author={{01.AI} and Young, Alex and Chen, Bei and Li, Chao and Huang, Chengen and Zhang, Ge and others},
  journal={arXiv preprint arXiv:2403.04652},
  year={2024}
}

@inproceedings{bai2024longalign,
  title={LongAlign: A Recipe for Long Context Alignment of Large Language Models},
  author={Bai, Yushi and Lv, Xin and Zhang, Jiajie and He, Yuze and Qi, Ji and Hou, Lei and Tang, Jie and Dong, Yuxiao and Li, Juanzi},
  booktitle={Findings of the Association for Computational Linguistics: EMNLP 2024},
  pages={1376--1395},
  year={2024}
}

@inproceedings{welbl2017sciq,
  title={Crowdsourcing Multiple Choice Science Questions},
  author={Welbl, Johannes and Liu, Nelson F. and Gardner, Matt},
  booktitle={Proceedings of the 3rd Workshop on Noisy User-generated Text},
  pages={94--106},
  year={2017}
}

@article{wang2024mmlupro,
  title={MMLU-Pro: A More Robust and Challenging Multi-Task Language Understanding Benchmark},
  author={Wang, Yubo and Ma, Xueguang and Zhang, Ge and Ni, Yuansheng and Chandra, Abhranil and Guo, Shiguang and Ren, Weiming and Arulraj, Aaran and He, Xuan and Jiang, Ziyan and others},
  journal={arXiv preprint arXiv:2406.01574},
  year={2024}
}

@article{taghanaki2024mmluproplus,
  title={MMLU-Pro+: Evaluating Higher-Order Reasoning and Shortcut Learning in LLMs},
  author={Taghanaki, Saeid Asgari and Khani, Aliasgahr and Khasahmadi, Amir},
  journal={arXiv preprint arXiv:2409.02257},
  year={2024}
}

@article{bertsch2025oolong,
  title={Oolong: Evaluating Long Context Reasoning and Aggregation Capabilities},
  author={Bertsch, Amanda and Pratapa, Adithya and Mitamura, Teruko and Neubig, Graham and Gormley, Matthew R.},
  journal={arXiv preprint arXiv:2511.02817},
  year={2025}
}

@inproceedings{mihaylov2018openbookqa,
  title={Can a Suit of Armor Conduct Electricity? A New Dataset for Open Book Question Answering},
  author={Mihaylov, Todor and Clark, Peter and Khot, Tushar and Sabharwal, Ashish},
  booktitle={Proceedings of the 2018 Conference on Empirical Methods in Natural Language Processing},
  pages={2381--2391},
  year={2018}
}
